\documentclass[times,twocolumn,final,authoryear]{elsarticle}

\usepackage{prletters}
\usepackage{framed,multirow}

\usepackage{amsmath}
\usepackage{amssymb}
\usepackage{latexsym}
\usepackage{algorithm}
\usepackage{url}
\usepackage[noend]{algpseudocode}
\usepackage[T1]{fontenc}
\usepackage{stmaryrd}
\usepackage{listings}
\usepackage{color} 
\usepackage[dvipsnames]{xcolor}
\usepackage{booktabs}
\usepackage{array} 
\usepackage{tabularx}
\usepackage[ruled,vlined,algo2e]{algorithm2e} 
\usepackage{multirow}
\usepackage{longtable}
\setcitestyle{square}

\definecolor{newcolor}{rgb}{.8,.349,.1}
\definecolor{mygreen}{RGB}{28,172,0} 
\definecolor{mylilas}{RGB}{170,55,241}

\definecolor{liamblue}{RGB}{0,150,255}

\journal{Pattern Recognition Letters}

\begin{document}

	\ifpreprint
	\setcounter{page}{1}
	\else
	\setcounter{page}{1}
	\fi
	
	\begin{frontmatter}
		
		\title{Leveraging Recent Advances in Deep Learning for Audio-Visual Emotion Recognition}
		
		\author[1]{Liam \snm{Schoneveld}} 
		\author[2]{Alice \snm{Othmani}\corref{cor1}} \cortext[cor1]{Corresponding author:  }
		\ead{alice.othmani@u-pec.fr}
		\author[2]{Hazem \snm{Abdelkawy}}

		\address[1]{Powder AI Research}
		\address[2]{Universit\'e Paris-Est, LISSI, UPEC, 94400 Vitry sur Seine, France}
		\received{1 May 2013}
		\finalform{10 May 2013}
		\accepted{13 May 2013}
		\availableonline{15 May 2013}
		\communicated{S. Sarkar}

		\begin{abstract}
			Emotional expressions are the behaviors that communicate our emotional state or attitude to others. They are expressed through verbal and non-verbal communication. Complex human behavior can be understood by studying physical features from multiple modalities; mainly facial, vocal and physical gestures. Recently, spontaneous multi-modal emotion recognition has been extensively studied for human behavior analysis. In this paper, we propose a new deep learning-based approach for audio-visual emotion recognition. Our approach leverages recent advances in deep learning like knowledge distillation and high-performing deep architectures. The deep feature representations of the audio and visual modalities are fused based on a model-level fusion strategy. A recurrent neural network is then used to capture the temporal dynamics. Our proposed approach substantially outperforms state-of-the-art approaches in predicting valence on the RECOLA dataset. Moreover, our proposed visual facial expression feature extraction network outperforms state-of-the-art results on the AffectNet and Google Facial Expression Comparison datasets.
		\end{abstract}
		
		
	\end{frontmatter}
	
	
	\section{Introduction}
	\label{introduction}
	
	Darwin concluded through his observations and descriptions of human emotional expressions that emotions adapt to evolution, are biologically innate, and universal across all human and even non-human primates (\cite{Matsumoto2001}). Formal, systematic research studies have since been realized on the universality of emotions. This work demonstrated: (i) the universality of six basic emotions (anger, disgust, fear, happiness, sadness and surprise) and (ii) the cultural differences in spontaneous emotional expressions (\cite{Ekman1987}). 
	
	A human's emotion resulting from an interaction with stimuli is referred to as an \textit{affect}. In psychology, an affect refers to the mental counterparts of internal bodily representations associated with emotions. In fact, humans express affect through facial, vocal or gestural behaviors. The notion of affect is subjective, and in the literature it is represented by two alternative views: \textbf{the categorical view} where affects are represented as discrete states with a wide variety of affective displays and \textbf{the dimensional view}, where we suppose that affects might not be culturally universal and alternatively, should be represented in a continuous arousal-valence space.
	
	Recently, a trend in the scientific community has emerged towards developing new technologies for processing, interpreting or simulating human emotions through Affective Computing or through Artificial Emotional Intelligence. Consequently, a broad range of applications have been developed in Human-Computer Interaction, health informatics and assistive technologies. Initial research on affect recognition focused mainly on unimodal approaches, with speech emotion recognition (SER) and facial expression recognition (FER) (\cite{RouastSurvey}) treated as separate problems. More recently however, work in affective computing has paid more attention to multimodal emotion recognition by developing approaches to multimodal data fusion.
	
	Research on affect recognition has seen considerable progress as the focus has shifted from the study of laboratory-controlled databases to databases covering real-world scenarios. In traditional emotion recognition databases, subjects posed a particular basic emotion in laboratory-controlled conditions. In more recent databases, videos are obtained from real-life scenarios with \textit{in-the-wild} environmental conditions and less constrained settings, which exhibit characteristics like illumination variation, noise, occlusion, non-frontal head poses, and so on. Today, automatic emotion recognition of the six basic emotions in acted visual and/or audio expressions can be performed with high accuracy. However, in-the-wild emotion recognition is a more challenging problem due to the fact that spontaneously occurring behavior varies more widely in its audio profile, visual aspects, and timing.
	
	In the present era, deep learning-based approaches are revolutionizing many areas of technology. Automatic emotion recognition likewise can benefit from the effectiveness of deep learning. In this paper, we propose a new approach for audio-visual emotion recognition (AVER). This approach is based on pre-training separate audio and visual deep convolutional neural network (CNN) recognition modules. A fusion module is then trained on the specific audio-visual dataset of interest. The fusion module is trained with the combination of generic emotion recognition features extracted by our pre-trained audio and visual components. The remainder of the paper is organized as follows. Section~\ref{sec:related_work} reviews the literature on AVER and presents the contributions of our paper. Section~\ref{sec:method} describes the proposed approach. Section~\ref{sec:results} presents the experiments then reports and discusses the results. Finally, Section~\ref{sec:conclusion} concludes the paper and suggests future work.
	
	\section{Related Works and paper contributions}
	\label{sec:related_work}
	
	\subsection{Related Work}
	Multimodal fusion for emotion recognition concerns the family of machine learning approaches that integrate information from multiple modalities in order to predict an outcome measure. Such is usually either a class with a discrete value (e.g., happy vs. sad), or a continuous value (e.g., the level of arousal/valence).  Several literature review papers survey existing approaches for multimodal emotion recognition (\cite{RouastSurvey, Baltrusaitis2018, Zeng2008,Poria2017}). There are three key aspects to any multimodal fusion approach: (i) which features to extract, (ii) how to fuse the features, and (iii) how to capture the temporal dynamics.\\
	\textbf{Extracted features:}  several handcrafted features have been designed for AVER. These low-level descriptors concern mainly geometric features like facial landmarks. Meanwhile, commonly-used audio signal features include spectral, cepstral, prosodic, and voice quality features.
	Recently, deep neural network-based features have become more popular for AVER. These deep learning-based approaches fall into two main categories. In the first, several handcrafted features are extracted from the video and audio signals and then fed to the deep neural network (\cite{Ringeval2015, He2015, Rejaibi2019, rejaibi2019mfcc, Muzammel2020}). In the second category, raw visual and audio signals are fed to the deep network (\cite{Tzarakis2017, Tzirakis2018, Basnet2019}). Deep convolutional neural networks (CNNs) have been observed as outperforming other AVER methods (\cite{RouastSurvey}).\\
	\textbf{Multimodal features fusion:}  An important consideration in multimodal emotion recognition concerns the way in which the audio and visual features are fused together. Four types of strategy are reported in the literature: feature-level fusion, decision-level fusion, hybrid fusion and model-level fusion (\cite{Zhang2017, Poria2017}). \\ \textit{Feature-level fusion} also called \textit{early-fusion} concerns approaches where features are immediately integrated after extraction via simple concatenation into a single high-dimensional feature vector. Such is the most common strategy for multimodal emotion recognition. \textit{Decision-level fusion} or \textit{late fusion} concerns approaches that perform fusion after an independent prediction is made by a separate model for each modality. In the audio-visual case, this typically means taking the predictions from an audio-only model, and the prediction from a visual-only model, and applying an algebraic combination rule of the multiple predicted class labels such as 'min', 'sum', and so on. \textit{Score-level fusion} is a subfamily of the decision-level family that employs an equally weighted summation of the individual unimodal predictors. \textit{Hybrid fusion} combines outputs from early fusion and from individual classification scores of each modality. 
	\textit{Model-level fusion} aims to learn a joint representation of the multiple input modalities by first concatenating the input feature representations, and then passing these through a model that computes a learned, internal representation prior to making its prediction. In this family of approaches, multiple kernel learning (\cite{chen2014}), and graphical models (\cite{Baltrusaitis2018, Baltrusaitis2013}) have been studied, in addition to neural network-based approaches.\\
	\textbf{Modelling temporal dynamics:} audio-visual data represents a dynamic set of signals across both spatial and temporal dimensions. \cite{RouastSurvey} identify three distinct methods by which deep learning is typically used to model these signals: \textit{Spatial feature representations:} concerns learning features from individual images or very short image sequences, or from short periods of audio. 
	\textit{Temporal feature representations:}  where sequences of audio or image inputs serve as the model's input. It has been demonstrated that deep neural networks and especially recurrent neural networks are capable of capturing the temporal dynamics of such sequences (\cite{Kim2017}). 
	\textit{Joint feature representations:} in these approaches, the features from unimodal approaches are combined. Once features are extracted from multiple modalities at multiple time points, they are fused using one of strategies of modality fusion (\cite{Ringeval2015}).

	\subsection{Contributions of this work}
	
	In this work, a higher-performing deep neural network-based approach for AVER is presented. The proposed model is a fusion of two deep neural networks: (i) a deep CNN model, trained with knowledge distillation, for FER and (ii) a modified and fine-tuned VGGish model for SER. A model-level fusion based approach is employed to fuse the audio and the visual feature representations. To model the temporal dynamics, the spatial and temporal representations are processed using recurrent neural networks. 
	The contributions of this work can be summarized as follows:
	\begin{itemize}
		\item A new high-performing deep neural network-based approach for AudioVisual Emotion Recognition (AVER)
		\item Learning two independent feature extractors -- one for audio and one for face images -- that are specialised for emotion recognition, and that could be employed for any downstream audiovisual  emotion  recognition  task or dataset
		\item Applying knowledge distillation (specifically, \emph{self-distillation}), alongside additional unlabeled data for FER
		\item Learning the spatio-temporal dynamics via a recurrent neural network for AVER.
	\end{itemize}

	\begin{figure*}
		\includegraphics[width=.97\textwidth]{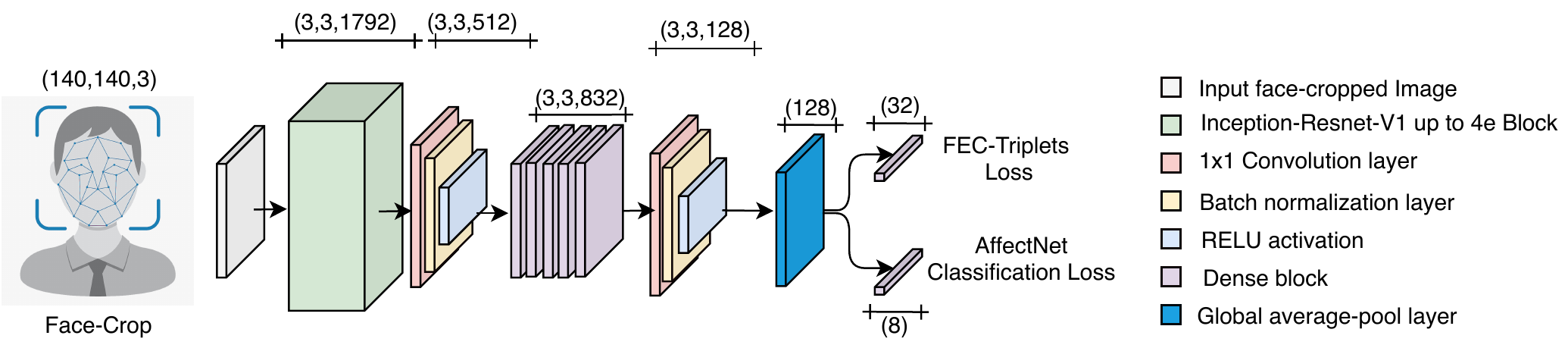}
		\caption{Our facial expression recognition neural network architecture, before distillation (i.e. the \emph{teacher network}). Faces are detected and cropped using MTCNN. The resulting 140x140 RGB images are then fed to an Inception Resnet V1 until the Inception 4e block. This is followed by a 1x1 convolution layer (1x1 Conv), batch normalization (BN) and a ReLU activation. Then, five DenseNet blocks are applied. Finally, another set of 1x1 Conv, BN and ReLU is applied. The output is then averaged over the spatial dimensions, resulting in a vector of size $D_{\text{face}}$ (in the figure, $D_{\text{face}}=128$). Two separate linear layers then give us the final model outputs -- a vector for the Google FEC triplets task, and class logits for AffectNet. The model is trained to minimize both the AffectNet and Google FEC losses simultaneously. The numbers over each block represent the tensor's output shape after applying that block.}
		\label{fig:fec_archi}
	\end{figure*}
	
	\begin{figure*}
		\includegraphics{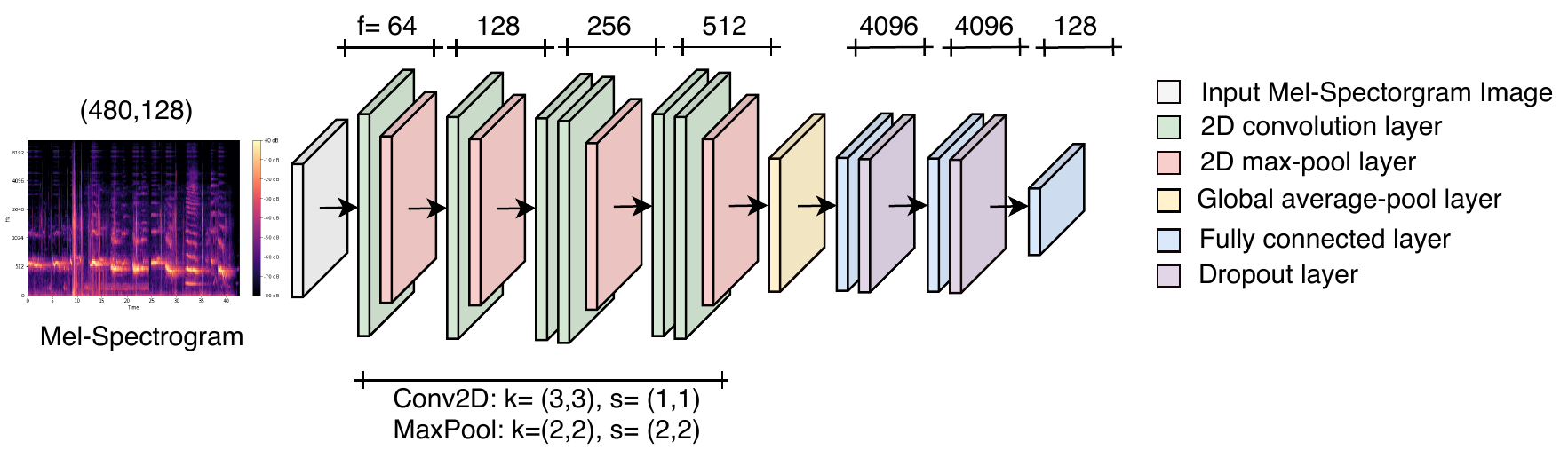}
		\caption{Modified VGGish backbone feature extractor for Speech Emotion Recognition. The Mel-Spectogram is computed from the audio signal and then fed to the modified VGGish backbone network consisting of 6 convolutional layers followed by 3 fully connected layers of size (4096, 4096 and 128) to output an embedding vector of size 128. The size of the feature maps (\textit{f}) of each convolutional and fully-connected layer are shown above each block of operations. The kernel size (\textit{k}) and stride (\textit{s}) are specified below the convolution blocks.}
		\label{fig:vggishModel}
	\end{figure*}
	
	\begin{figure*}
		\includegraphics[width=.97\textwidth]{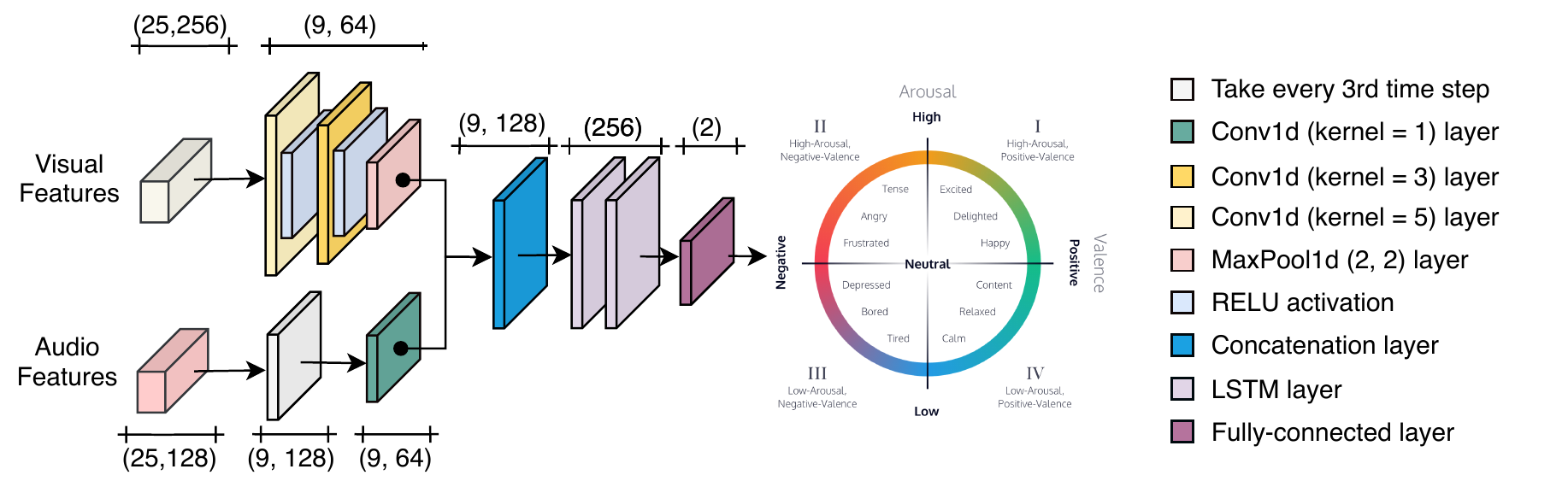}
		\caption{Our audio-visual fusion network architecture. The audio and the visual embedding vectors are each fed to a small, independent convolutional network. This results in one tensor of size (9, 64) for each modality. We concatenate these two to give a tensor of size (9, 128), which is fed to a two-layer LSTM network with dimensionality of 256. Taking the final time step's output of this LSTM gives a single vector of size 256, which is pass through a single fully-connected layer with two outputs, which after a tanh activation gives our predictions between -1 and 1 for arousal and valance.}
		\label{fig:fusionarchi}
	\end{figure*}
	
	\section{Proposed multimodal deep CNN architecture}
	\label{sec:method}
	
	Our proposed multimodal deep CNN architecture is made up of three components:
	
	\subsection{Visual facial expression embedding network}
	\label{sec:cnn}
	
	The first component of our multimodal architecture is a deep convolutional neural network (CNN) for facial expression recognition. The input to this network is a single RGB face image, detected and cropped using Multi-Task Cascaded Convolutional Networks (MTCNN) (\cite{MTCNN}). The output of this network is a compact vector of dimension $D_{\text{face}}$.
	
	We refer to this network as our `facial expression embedding network', and we train it using \textit{knowledge distillation} (\cite{hinton2015}). Knowledge distillation is a two step process whereby a \textit{teacher} network is trained on the task of interest, and then a (typically smaller) \textit{student} network is trained on predictions made by the teacher. Specifically in this work, we leverage the benefits of \textit{self-distillation}, whereby the student network has the same size as (or at least, is not smaller than) the teacher network. Self-distillation was recently leveraged to achieve state-of-the-art results on the well-known Imagenet classification dataset (\cite{xie2020self}). It has also been shown theoretically that self-distillation can improve a model's performance via a regularization effect (\cite{mobahi2020self}). We use self-distillation to improve the performance of our facial expression embedding network. The training procedure for this network thus consists of two phases:
	\begin{enumerate}
		\item Training a teacher model: our teacher model is a fine-tuned FaceNet (\cite{Schroff2015}), trained simultaneously on two different visual facial expression recognition datasets (Section \ref{sec:teacher}).
		\item Training a student model: a second CNN is additionally trained to mimic the outputs of this fine-tuned FaceNet (Section \ref{sec:student}).
	\end{enumerate}
	
	\subsubsection{The teacher network}
	\label{sec:teacher}
	
	The starting point for our teacher model is a pre-trained FaceNet (\cite{Schroff2015}). This model is then trained to learn specialised features for facial emotion recognition using two datasets:
	\begin{itemize}
		\item \textbf{AffectNet} (\cite{AffectNet}), which consists of around 440,000 in-the-wild face crop images, each of which is human-annotated into one of eight facial expression categories (Neutral, Happy, Sad, Surprise, Fear, Disgust, Anger and Contempt).
		\item \textbf{Google Facial Expression Comparison (FEC)} (\cite{GoogleFEC}), which consists of around 700,000 triplets of unique face crop images. Annotations denoting the most similar pair of face expressions in each triplet are provided. The goal is to train a model that places the similar pair closer together in a learned embedding space.
	\end{itemize}
	
	Our teacher model's architecture (Fig. \ref{fig:fec_archi}) is almost identical to the model proposed in \cite{GoogleFEC}, the only difference is that we add an additional output head for the AffectNet loss. A pre-trained FaceNet\footnote{In this, we used a FaceNet pre-trained on the VGGFace2 dataset, as we found performance to be slightly improved. The pre-trained FaceNet model architecture and weights were obtained from https://github.com/timesler/facenet-pytorch.} is taken up until the Inception 4e block. This is followed by a 1x1 convolution and a series of five untrained DenseNet (\cite{DenseNet}) blocks. After this, another 1x1 convolution followed by global average pooling reduces this representation to a single $D_{\text{face}}$ dimensional vector. After pooling, two independent linear transformations serve as output heads. These heads take the $D_{\text{face}}$-dimensional facial expression representation vector as input and make separate predictions for the AffectNet and FEC tasks. A 32-dimensional embedding is used for the FEC triplets task, while an 8-dimensional output head produces class logits for AffectNet (which has 8 classes). The teacher network training procedure is detailed in Algorithm \ref{algo1}, while implementation details are provided in the supplementary materials.
	
	To improve the regularization effects of self-distillation through model ensembling, we in fact train two teacher networks, and concatenate their outputs to serve as distillation targets (see Section \ref{sec:student} for details). The only difference between the two teachers are the random seeds used for initialization, and penultimate layer dimensionalities: we use $D_{\text{face}}=128$ for one teacher network, and $D_{\text{face}}=256$ for the other.
	
	\subsubsection{Student network}
	\label{sec:student}
	
	Our student network is a DenseNet201 pre-trained on Imagenet.\footnote{We use the implementation and pre-trained Imagenet weights provided in the \texttt{torchvision} Python package.} The student network training procedure is essentially the same as described in Algorithm \ref{algo1}, except that we additionally sample batches of unlabeled data from an internal dataset, which we refer to as \textit{PowderFaces}. The PowderFaces dataset was created by downloading approximately 20,000 short, publicly-available videos from various online sources such as YouTube. To increase the frequency of faces in the dataset, specific search terms and topics were used when searching for videos, such as `podcast', `interview', or `monologue'. MTCNN face detection was then applied to the extracted frames from those videos, producing approximately 1 million individual face crops. The sampled batches of face crops from the Google FEC, AffectNet, and PowderFaces datasets are passed through our two teacher networks. Each of the two teacher networks produces predictions for the Google FEC task (32-dimensional) and AffectNet class logits (8-dimensional). These four vectors (i.e., two vectors from two teacher networks) are individually L2-normalized The four normalized vectors are then concatenated, producing one long vector of dimension 80. A knowledge distillation loss (we specifically use `Relational Knowledge Distillation' (\cite{park2019relational}) for our loss function) is then calculated by comparing the output of a third output head in the student network to this 80-dimensional target vector. This knowledge distillation loss is then added to the standard AffectNet and Google FEC losses, which are calculated as per the teacher network training procedure. Implementation details for our student network are provided in the supplementary materials.
	
	\subsection{Audio embedding network for emotion recognition}
	
	This section details our proposed deep learning-based approach for recognizing emotions from audio segments. The approach is based on fine-tuning the VGGish model (\cite{Vggish2017}) on the RECOLA dataset (\cite{recola}).
	
	\subsubsection{Audio pre-processing}
	\label{sec:audiopreproc}
	
	For each input audio file, a set of Mel-spectrogram representations ($R$) are created. The input audio files are down-sampled at a 16KHz sampling rate. Then, the short-time Fourier transform (STFT) is performed to create windows with length ($l$) of 40 milliseconds and a hop length of 40 milliseconds. To create the latter windows, a set of 128 Mel filters ($M_f$) are applied with a Mel frequency range of 125-7500 Hz. Finally, for each audio file, a tensor of shape $[R,l,M_f]$ is generated to create a compatible pre-processed data input for the VGGish backbone network. 
	
	\begin{algorithm2e}
		
		\SetAlgoLined
		\For{iteration in range($N$)}{
			$(\mathbf{X}_{\text{FEC}}, \mathbf{y}_{\text{FEC}})$ $\leftarrow$ batch of Google FEC triplets and labels \\
			$(\mathbf{X}_{\text{Aff}}, \mathbf{y}_{\text{Aff}})$ $\leftarrow$ batch of AffectNet images and class labels\\
			$\mathbf{e}_{\text{FEC}} \leftarrow f_{\Theta}(\mathbf{X}_{\text{FEC}})$ \Comment{Face embeddings for FEC images}\\
			$\mathbf{e}_{\text{Aff}} \leftarrow f_{\Theta}(\mathbf{X}_{\text{Aff}})$ \Comment{Face embeddings for AffectNet images}\\
			$\mathbf{v}_{\text{FEC}} \leftarrow g_{\phi}(\mathbf{e}_{\text{FEC}})$ \Comment{Predict vectors for triplet loss}\\
			$\mathbf{p}_{\text{Aff}} \leftarrow h_{\theta}(\mathbf{e}_{\text{Aff}})$ \Comment{Predict class probabilities for AffectNet}\\
			$L_\text{FEC} = \texttt{triplet\_loss}(\mathbf{v}_{\text{FEC}},  \mathbf{y}_{\text{FEC}})$\\
			$L_\text{Aff} = \texttt{cross\_entropy\_loss}(\mathbf{p}_{\text{Aff}}, \mathbf{y}_\text{Aff})$\\
			$L = L_\text{FEC} + \alpha * L_\text{Aff}$ \Comment{Total loss for training step}\\
			Obtain all gradients $\Delta_\text{all} = (\frac{\partial L}{\partial \Theta}, \frac{\partial L}{\partial \phi}, \frac{\partial L}{\partial \theta})$\\
			$(\Theta, \phi, \theta) \leftarrow \texttt{SGD}(\Delta_\text{all}$) \Comment{Update feature extractor and output heads' parameters simultaneously}
		}
		\caption{Visual model: training the teacher network. Given feature extractor network $f_{\Theta}$, Google FEC output head $g_{\phi}$, AffectNet output head $h_{\theta}$,  number of training steps $N$, AffectNet loss weight $\alpha$.}
		\label{algo1}
	\end{algorithm2e}

	\subsubsection{VGGish backbone network}
	\label{sec:vggish}
	
	Our deep model for audio-based emotion recognition is based on a modified version of the VGGish model (\cite{Vggish2017}). Our starting point is the original VGGish model, pre-trained on the Audio Set dataset (\cite{audioSet}). The VGGish backbone consists of 6 convolutional layers that output 64, 128, 256, and 512 feature maps ($f$) respectively. For each convolution layer, a kernel ($k$) with size 3x3, and stride ($s$) of 1x1 is used. A max pooling layer with a kernel ($k$) of size 2x2, and stride ($s$) 2x2 is then applied. We take this VGGish backbone, but replace its last convolution and max pooling layers with a global average pooling layer. The resulting model produces an output vector of dimension $256$. After this, we add three randomly-initialized, fully-connected layers, with output dimensionalities of $4096$, $4096$, and $128$. These layer sizes were chosen to mimic the fully-connected penultimate layers of the original VGGish architecture. The aim of these layers is to extract a standard embedding vector with size of $128$ that reflects the emotional characteristics of the input audio segment.
	
	We take this expanded VGGish backbone architecture and fine-tune it end-to-end on the RECOLA dataset. Two separate VGGish networks are fine-tuned: one to predict arousal and the other to predict valence. We pass inputs of size $[480,128]$ to the VGGish model, which are the mel-spectrogram representations of 30 seconds of audio from one of the videos in the RECOLA dataset. The target used for fine tuning is then the average ground truth arousal or valence for the target values corresponding to the input 30 seconds of audio. We predict this target by passing the $128$-dimensional audio representation through a fully-connected layer $f_{\phi}$ with a \texttt{tanh} activation. The training procedure for fine tuning our audio feature extraction model is detailed in Algorithm \ref{algo:audio}.
	
	\begin{algorithm2e}
		\SetAlgoLined
		\For{iteration in range($N$)}{
			$(\mathbf{X}, \mathbf{y})$ $\leftarrow$ batch of RECOLA spectrograms and targets \\
			$\mathbf{e} \leftarrow f_{\Theta}(\mathbf{X})$ \Comment{Calculate VGGish embeddings for batch}\\
			$\mathbf{p} \leftarrow f_{\theta}(\mathbf{e})$ \Comment{Predict arousal for all elements in batch}\\
			$Loss = -\texttt{concordance\_correlation\_coeff}(\mathbf{p}, \mathbf{y})$\\
			Obtain all gradients $\Delta_\text{all} = (\frac{\partial Loss}{\partial \Theta}, \frac{\partial Loss}{\partial \theta})$\\
			$(\Theta, \theta) \leftarrow \texttt{Adam}(\Delta_\text{all}$) \Comment{Update VGGish model, output head}
		}
		\caption{VGGish fine-tuning algorithm for predicting arousal. Given the VGGish feature extractor network $f_{\Theta}$, arousal prediction head $f_{\phi}$, number of training steps $N$.}
		\label{algo:audio}
	\end{algorithm2e}
	
	\subsection{Audio-visual fusion model}
	\label{sec:fusion}
	
	A model-level fusion based approach is considered as shown in Fig. \ref{fig:fusionarchi}. Visual features are extracted by taking face crops from the video sequence of interest using MTCNN and passing them to our student network (Section \ref{sec:student}). Audio features are extracted using our fine-tuned VGGish backbone (the temporal granularity of these features is increased by removing the global average pooling over the temporal dimension from our fine-tuned VGGish model). To have the same reduced size, audio and visual features are passed through small, independent pre-transform networks consisting of 1D convolutions, where the convolution filters slide over the temporal dimension. The pre-transform networks are designed to give an output tensor of shape $[9, 64]$. These are concatenated into a single audio-visual features tensor, which is passed to a two-layer LSTM network with hidden and output dimensionality of $256$. The final output of the LSTM network is then taken, and passed to a simple linear transform with two outputs. These outputs are passed through a \texttt{tanh} activation to produce the final arousal and valence predictions. Our loss function is the negative CCC. Further implementation details for our fusion network are provided in the supplementary materials.
	
	\section{Experimental results}
	\label{sec:results}
	
	\subsection{Datasets}
	\label{sec:datasets}
	The performances of the proposed approach have been evaluated using the REmote COLlaborative and affective interactions (RECOLA) corpus (\cite{recola}). In RECOLA, participants' spontaneous interactions were collected while being engaged in a remote discussion that aimed to manipulate their moods. Then, six annotators measured the emotional state present in all sequences continuously on the valence and arousal dimensions. 27 audio-visual recordings of 5 minutes of interaction, which includes 9 videos for training and 9 for validation, are publicly available. In order to perform a fair comparison, the test set annotations (for the last 9 videos) of the AVEC challenge are not given. We use the training videos to train our models, validate on the validation videos, and submitted our results on the test set to the RECOLA dataset managers for evaluation. We also evaluate our visual feature extraction network on held-out evaluation data from the AffectNet and Google FEC datasets, which were introduced in Section \ref{sec:teacher}.
	
	\subsection{Evaluation metric}
	\label{sec:ccc}
	The Concordance Coefficient Correlation (CCC) (\cite{lawrence1989}) is used to evaluate the performance of the proposed approach on RECOLA, as it is standard metric for emotion recognition on the RECOLA dataset. The CCC (Equation \ref{eqn:ccc}) measures the agreement between a vector of predicted ($Pred$) and true ($True$) values for a continuous variable:
	
	\begin{equation}
	CCC(True,Pred)=\frac{2 * {Corr(True, Pred) } * {\sigma_{True}} * {\sigma_{Pred}}} {\sigma^2_{True}+\sigma^2_{Pred}+(\mu_{True}-\mu_{Pred})^2}
	\label{eqn:ccc}
	\end{equation}
	
	Where $\mu_{\mathbf{x}}$ is the mean of $\mathbf{x}$, $\sigma_{\mathbf{x}}$ is the standard deviation of $\mathbf{x}$, and $Corr(\mathbf{x}, \mathbf{y})$ returns Pearson's correlation coefficient between $\mathbf{x}$ and $\mathbf{y}$.
	
	\subsection{Visual facial expression embedding network performance}
	\label{sec:visperf}
	
	We evaluate our visual facial expression embedding network on the standard evaluation subsets of the two datasets it was trained on:
	\begin{enumerate}
		\item \textbf{AffectNet:} for AffectNet, which requires classifying faces into eight discrete facial expression classes, we train a logistic regression model on the features extracted by our student network for the entire AffectNet training set.\footnote{When training this logistic regression, we re-weight the classes in the AffectNet training set to have equal representation, as per the validation set.} This method achieves state-of-the-art results on the AffectNet validation set, with an accuracy of $61.6\%$ (Table \ref{tab:affectnet_results}).
		\item \textbf{Google FEC:} following \cite{GoogleFEC}, we evaluate using triplet accuracy on the Google FEC test set. Using this metric, we find our approach substantially improves on state-of-the-art on the FEC test set, with an accuracy of $86.5\%$ (Table \ref{tab:fec_results}).
	\end{enumerate}
	
	To experimentally verify the importance of the different components of our approach, we perform an ablation study. Training the student model architecture without the distillation loss component reveals the importance of distillation. Without distillation, our model's performance dropped substantially: to 58.8\% on AffectNet and 85\% on Google FEC.
	
	Similarly, to determine the importance of the unlabeled PowderFaces dataset, we again train the student model with distillation, but without the additional distillation targets provided by using this unlabeled data. The results suggest that the additional unlabeled data may not be so important to our results: accuracy on AffectNet dropped only slightly to 61.1\%, and performance on Google FEC reduced by only 0.1\% to 86.4\%. We discuss these ablation results further in Section \ref{sec:discussion}.
	
	\begin{table} 
		\caption{Performances of the proposed visual facial expression embedding network on the AffectNet validation set comparing to existing state-of-the-art methods}
		\begin{tabular*}{\columnwidth}{l|l} 
			\hline 
			\textbf{Methods} & \textbf{Accuracy}\\
			\hline
			\cite{Georgescu2019}& 59.6\%\\
			
			\cite{Siqueira2020}& 59.3\%\\
			
			Ours (Teacher model) & 61.3\%\\
			
			Ours (Student, no distillation) & 58.8\%\\
			
			Ours (Distilled student, no PowderFaces) & 61.1\%\\
			
			Ours (Distilled student) &  \textbf{61.6\%}\\
			\hline
		\end{tabular*}
		\label{tab:affectnet_results}
	\end{table}
	
	\begin{table}
		\caption{Triplet prediction performances of the proposed visual facial expression embedding network on the Google FEC test set comparing to existing state-of-the-art methods }
		\begin{tabular*}{\columnwidth}{l|l} 
			\hline
			\textbf{Methods} & \textbf{Accuracy} \\
			\hline
			\cite{GoogleFEC}& 81.8\% \\
			Ours (Teacher model) & 84.5\% \\
			Ours (Student, no distillation) & 85.0\% \\
			Ours (Distilled student, no PowderFaces) & 86.4\% \\
			Ours (Distilled student) & \textbf{86.5\%}\\
			\hline
		\end{tabular*}
		
		\label{tab:fec_results}
	\end{table}

	\subsection{Performance on RECOLA in the visual-only and audio-only context}
	To test performance on RECOLA of our visual and audio feature extractors separately, we retrain the same fusion architecture, but disable either the audio or visual feature inputs.\\

	\textbf{Visual-only:} feeding the embeddings from our visual feature extractor into the visual-only version of our fusion model performs well on the RECOLA dataset (Table \ref{tab:RECOLA_performances}). Such reaches a CCC of $0.55$ for predicting valence and $0.57$ for predicting arousal on the validation set, while on the test set our CCC reaches $0.66$ for valence and $0.57$ for arousal. This result illustrates the robustness of our visual feature extractor: when predicting valence, our method achieves state-of-the-art performance when compared to other \textit{multimodal} approaches, even though we use \textit{only visual} features as input.\\
	\textbf{Audio-only:} our results (Table \ref{tab:RECOLA_performances}) show that our modified VGGish backbone feature extractor for audio segments performs well, CCCs of $0.52$ and $0.70$ for valence and arousal, respectively on the RECOLA test set. The achieved results for arousal prediction match the existing state-of-the-art methods when only audio features are used (Table \ref{tab:vggish_results_comparison}). This shows that our approach to transfer learning from the acoustic events domain, and the VGGish architecture, provide a robust means to extracting audio-based features for emotion recognition.
	
	\subsection{Fusion model performance and comparison with state-of-the-art approaches}
	
	Our multimodal fusion model achieves a CCC of $0.740$ in valence prediction, and $0.719$ in arousal prediction in RECOLA test set (Table \ref{tab:comparison_existing_approaches}). These results reflect the robustness of our learned visual and audio features extraction techniques, and the efficacy of the approach to fusing these modalities and accounting for temporal dynamics. Our proposed method, with a CCC of $0.740$, substantially outperforms all existing methods in predicting valence, with the previous best performing approach of \cite{Tzarakis2017} achieving a CCC of $0.612$. Simultaneously, our approach achieves strong results in predicting arousal with a CCC of $0.719$, compared to the state-of-the-art performance of \cite{Ringeval2015}, who obtained a CCC of $0.796$.
	
	\begin{table}
		\caption{RECOLA dataset results (in terms of CCC) for predicting arousal and valence on train, development and test sets.}
		\begin{tabular}{l|lll|lll}
			\hline
			& \textbf{Valence}                    &                          &      & \textbf{Arousal}                    &                          &      \\ \hline
			\textbf{CCC}  & \multicolumn{1}{l|}{Train} & \multicolumn{1}{l|}{Dev} & Test & \multicolumn{1}{l|}{Train} & \multicolumn{1}{l|}{Dev} & Test \\ \hline
			Visual only & \multicolumn{1}{l|}{.6}    & \multicolumn{1}{l|}{.55} & .66  & \multicolumn{1}{l|}{.49}   & \multicolumn{1}{l|}{.57} & .57  \\ \hline
			Audio only  & \multicolumn{1}{l|}{.55}   & \multicolumn{1}{l|}{.46} & .52  & \multicolumn{1}{l|}{.78}   & \multicolumn{1}{l|}{.80}  & .70   \\ \hline
			Audio-visual & \multicolumn{1}{l|}{.69}   & \multicolumn{1}{l|}{.63} & .74  & \multicolumn{1}{l|}{.78}   & \multicolumn{1}{l|}{.81} & .72  \\ \hline
		\end{tabular}
		\label{tab:RECOLA_performances}
	\end{table}
	
	\begin{table}
		\caption{Performances of the proposed audio embedding network on the RECOLA dataset comparing to existing state-of-the-art methods. In parenthesis are the performances obtained in the development set. ------ : no results reported in the original papers.}
		\begin{tabular*}{\columnwidth}{l|c|r}
			\hline
			\textbf{Methods} & \textbf{Arousal} & \textbf{Valence}  \\
			\hline
			\cite{Tzarakis2017}& .70 (.75) & .31 (.41) \\
			\cite{Han2017} & .67 (.76) & .36 (.48) \\
			\cite{He2015} &------(.80) &------(.40) \\
			Ours & .70 (.80) & .52 (.46) \\
			\hline
		\end{tabular*}
		\label{tab:vggish_results_comparison}
	\end{table}
	
	\begin{table*}
		\caption{RECOLA dataset results (in terms of CCC) for predicting arousal and valence. S.M.: Strength modeling of SVR + BLSM }
		\begin{tabular}{l|l|l | l |c|r } 
			\toprule 
			\textbf{Methods} & \textbf{Audio features} & \textbf{Visual features} & \textbf{Modality fusion} &\textbf{Arousal} & \textbf{ Valence}\\
			\midrule
			\midrule
			\cite{Ringeval2015} & LLDs + BLSTM & LLDs & Feature-level & .761 & .492\\  
			\midrule
			\cite{Ringeval2015} & LLDs + BLSTM& LLDs & Decision-level & .796 & .501 \\
			\midrule
			\cite{He2015} & LLDs & LLDs & model-based (BLSTM) & .747 & .609 \\
			\midrule
			\cite{Han2017} & LDDs+S.M.  & Geom. + S.M.  & modality and model-based & .685 & .554\\
			\midrule
			\cite{Tzarakis2017}& 1D CNN & ResNet50  & Model-level (2 LSTM) & .714 & .612  \\
			
			\midrule
			
			\textbf{Proposed} & \textbf{Fine-tuned VGGish} & \textbf{Distilled CNN} & \textbf{Model-based (LSTM)}  & \textbf{.719} & \textbf{.740}\\
			\bottomrule
		\end{tabular}
		
		\label{tab:comparison_existing_approaches}
	\end{table*}
	
	\subsection{Discussion}
	\label{sec:discussion}
	
	The results in Section \ref{sec:visperf} suggest that the unlabeled PowderFaces dataset only provides a marginal benefit in performance, if any at all. This ran contrary to our expectations: we expected unlabeled data to help in this context, given a similar approach achieved state-of-the-art results on Imagenet (\cite{xie2020self}). In that work however, the unlabeled dataset consisted of 300 million images -- 300 times more than the Imagenet dataset itself. In our case, our unlabeled dataset of one million images is only about twice the size of the number of faces in AffectNet and Google FEC combined. Thus, to truly conclude whether the benefits of unlabeled data as shown on Imagenet transfer well to facial expressions, many more unlabeled images are needed. We plan to address this point in future work.
	
	On the other hand, it appears that self-distillation is indeed beneficial in the facial expressions domain. This is illustrated by the marked improvement in our accuracy on the Google FEC dataset when distillation is applied, compared to training without distillation. For AffectNet, the benefits of distillation are less pronounced, and it seems that the combination of using a pre-trained FaceNet model, plus training on Google FEC at the same time as AffectNet are all necessary components to achieving our results. 
	
	Finally, our results for valence on the RECOLA test set show a dramatic improvement over existing state-of-the-art approaches (Table \ref{tab:comparison_existing_approaches}). As mentioned, our visual-only model for RECOLA also achieved state-of-the-art for valence (Table \ref{tab:RECOLA_performances}). This provides further evidence for the effectiveness of our visual feature extraction approach.
	
	\section{Conclusion and Future work}
	\label{sec:conclusion}
	This paper introduces a high-performing deep neural network-based approach for AVER that fuses a distilled visual feature extractor network with a modified VGGish backbone and a model-level fusion architecture. The proposed visual facial expression embedding network shows that end-to-end training on both AffectNet and FEC in tandem is a highly effective method for learning robust facial expression representations. We have also demonstrated that knowledge distillation can provide further improvements for facial expression recognition. The performance of our modified VGGish backbone feature extractor presents a promising new direction for predicting emotion from audio. Moreover, our deep neural network approach to multimodal fusion has been shown to be effective in AVER, outperforming the state-of-art methods in predicting valence on the RECOLA dataset. For future work, we plan to investigate the best strategy for continuous emotion encoding: classification with coarse categories, regression, label distribution learning or even ranking.
	
	\section*{Acknowledgements}
	This work was supported by Powder, a deep tech startup. Powder is a video editing and sharing platform for gamers. \url{https://powder.gg/}
	
	\section*{Supplementary Materials}
	\label{sec:supp}
	
	Supplementary material associated with this article can be found in the enclosed file.
	
	\bibliographystyle{model2-names}

	\bibliography{refs}
	
\end{document}